\documentclass[conference]{IEEEtran}

\usepackage{graphicx}
\usepackage{amsmath,amssymb,bm}
\usepackage{booktabs}
\usepackage{array}
\usepackage{siunitx}
\usepackage{cite}
\usepackage{url}
\usepackage{xcolor}
\usepackage{balance}
\usepackage{placeins}
\usepackage{multirow}

\newcommand{\MotherMass}{14.0}
\newcommand{\ChildMass}{1.9}
\newcommand{\DeckDia}{0.95}
\newcommand{\ChildDia}{0.75}

\title{RTK–Vision PPO for Autonomous Micro UAV Recovery on an Airborne Carrier}
\author{Anonymous Authors}

\author{\IEEEauthorblockN{Aashish Sahu}
\IEEEauthorblockA{\textit{Department of Mechanical and Aerospace Engineering} \\
\textit{Indian Institute of Technology Hyderabad}\\
Kandi, Sangareddy, India \\
me22resch11011@iith.ac.in}
\and
\IEEEauthorblockN{R Prasanth Kumar}
\IEEEauthorblockA{\textit{Department of Artificial Intelligence} \\
\textit{Indian Institute of Technology Hyderabad}\\
Kandi, Sangareddy, India \\
rpkumar@ai.iith.ac.in}
}

\begin{document}
\maketitle

\begin{abstract}
Autonomous recovery of a micro unmanned aerial vehicle (UAV) onto a moving airborne carrier enables reusable deploy--mission--recover operation, but couples long-range rendezvous, close-range perception, carrier motion, aerodynamic interaction, and a discontinuous contact event. This paper presents an RTK--vision-guided reinforcement-learning framework in which a child UAV is physically transported by a larger carrier, takes off from the carrier while airborne, executes an independent sortie, returns to the carrier's \emph{current} position, redocks, and subsequently descends with the carrier. Both vehicles carry RTK-GNSS, and the carrier continuously shares its navigation state with the child. Near the recovery deck, RTK remains active while a downward-facing camera with a fiducial marker detector provides marker-relative alignment cues. A proximal policy optimization (PPO) policy governing the terminal recovery phase is trained in a physics-based MuJoCo simulation environment with explicit sensor noise models, an aerodynamic disturbance surrogate, and marker-latency randomization, then transferred to hardware; PX4 retains low-level stabilization and a deterministic safety gate authorizes descent independently of the learned policy. The PPO checkpoint achieves 99.55\% success over 2,000 held-out randomized terminal episodes, compared to 78.4\% for a tuned PD baseline under identical conditions, with a median planar terminal error of \SI{6.62}{cm}. Across 14 outdoor trials the full mission succeeds in 13 (\SI{92.9}{\percent}), spanning both near-region recovery and recovery after the carrier translates away from the release point. The results demonstrate a complete autonomous aerial deployment-and-recovery cycle rather than an isolated landing maneuver, establishing a practical basis for reusable carrier--child operation in inspection, surveillance, and mobile-logistics applications.
\end{abstract}

\section{Introduction}

Small UAVs are attractive for inspection, environmental monitoring, inventory observation, search-and-rescue, and localized sensing because their low mass and compact geometry permit operation close to structures and in constrained spaces. Their usefulness is nevertheless limited by endurance: battery capacity sufficient for long-distance transit increases mass and reduces the agility that motivates use of a micro UAV. A carrier--child architecture separates transport from task execution. A larger carrier performs the energy-intensive transit and loitering segment, deploys a smaller child near the region of interest, and later recovers it for reuse.

\begin{figure}[t]
    \centering
    \includegraphics[width=\linewidth]{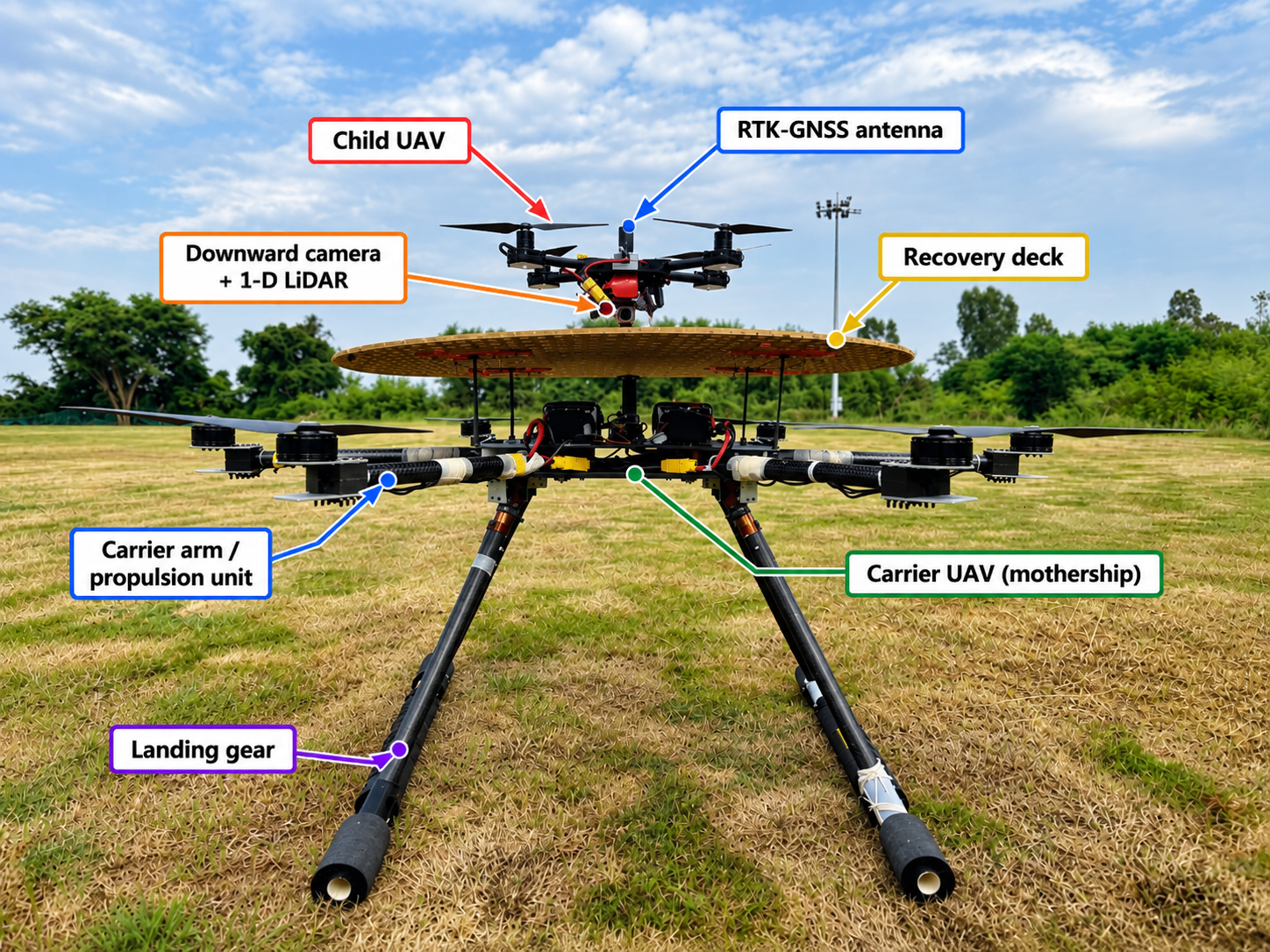}
    \caption{The mother--child aerial recovery platform used in outdoor experiments, showing the child quadrotor on the carrier recovery deck with RTK-GNSS, downward-facing camera, and 1-D LiDAR.}
    \label{fig:placeholder}
\end{figure}

Closing this cycle is substantially harder than an air launch. The child must return to a target that is itself a flying robot, has residual position and attitude motion, generates a rotor wake, and must support the recovered payload after contact. Moving-ground-platform landing has been studied using visual servoing, prediction, and disturbance-aware control \cite{lee2012,falanga2017,borowczyk2017,baca2019autonomous,paris2020}, but the target in those studies is ground-supported. Direct aerial landing is less mature. Wang \emph{et al.} demonstrated autonomous takeoff and landing of a mission multirotor on a larger six-rotor platform at centimeter-level accuracy using vision and an RBF-PID controller~\cite{wang2022dual}. Shen \emph{et al.} formulated constrained convex trajectory planning for landing on an aerial vehicle carrier, reporting feasible trajectories within safety constraints~\cite{shen2024convex}. Dong \emph{et al.} modeled carrier-induced aerodynamic interference and demonstrated landing on both hovering and translating aerial platforms with approximately \SI{13}{cm} dynamic accuracy over three trials~\cite{dong2024aerial}, though both vehicle poses were supplied by an external OptiTrack system and visual navigation was identified as a future extension. In the terminal phase of our system, a downward-facing camera detects a fiducial marker affixed to the recovery deck; engineered fiducials—including ArUco patterns, AprilTags, and QR-code variants—are widely employed to reduce detection ambiguity and provide pose cues during the final meters of precision landing \cite{olson2011apriltag,garrido2014aruco,kalaitzakis2021fiducial,baca2019autonomous}, and have been applied to UAV landing benchmarks \cite{semerikov2025vision} and docking on ground platforms \cite{lange2009autonomous}.

The problem addressed here is broader than terminal landing. The child begins physically seated on the mothership, takes off directly from the recovery deck while the carrier is airborne, performs an independent sortie, reacquires the \emph{same} carrier, lands back on it, and is returned toward the ground as a coupled stack. Long-range recovery uses continuously shared RTK-GNSS rather than a stored launch coordinate, and onboard vision is used concurrently with RTK for terminal alignment. The PPO policy governs only the terminal recovery phase—outputting the child's three-axis carrier-relative velocity reference from marker acquisition to capture—while PX4 retains inner-loop stabilization and a deterministic safety gate provides a policy-independent layer of descent authorization; the carrier is not controlled by the policy.

\textbf{Contributions.} This work delivers an integrated autonomous \emph{deploy--mission--recover} system with contributions at four levels. \emph{(1)~Systems and experimental:} we demonstrate a complete airborne deploy--mission--recover cycle in which the same child launches from the flying carrier, executes an independent sortie, and is recovered onto the same carrier before coupled descent—validated outdoors over 14 trials with fully onboard sensing and no external motion capture, unlike prior aerial-carrier work addressing only terminal landing. \emph{(2)~Autonomy integration:} recovery targets the carrier's continuously shared, \emph{current} RTK state rather than a memorized launch coordinate, with overlapping RTK and fiducial vision in which observation staleness ($s_V,\tau_V$) is made explicit to the policy, and dual independent onboard docking confirmation ($q_{\rm dock}=q_C\wedge q_M$) before descent. \emph{(3)~Safety formulation:} we introduce a deterministic safety-gate layer decoupled from the learned policy that authorizes descent only under lateral-alignment, freshness, tilt, and speed conditions, with abort-and-retry, bounding the consequence of policy uncertainty near the deck. \emph{(4)~Learning and evaluation:} the policy is trained in a MuJoCo environment with characterized sensor noise, marker latency, and an aerodynamic-wake surrogate, then transferred to hardware, with a 2,000-episode evaluation (PPO 99.55\% vs.\ tuned PD 78.4\%; \SI{6.62}{cm} median error), two complementary outdoor conditions, and a flight-log failure-mode analysis. The two outdoor conditions directly test the operational requirement that the child return to the carrier's current state rather than a memorized release point.

\section{Related Work}
\subsection{Moving-platform landing and aerial recovery}

Vision-based landing on moving ground targets has been demonstrated using image-based servoing~\cite{lee2012}, target-state prediction~\cite{falanga2017,borowczyk2017}, and onboard sensing~\cite{baca2019autonomous}. Paris \emph{et al.} additionally studied dynamic landing in turbulent wind~\cite{paris2020}. These works establish important relative-navigation and velocity-matching principles, but do not address mutual rotor interaction or airborne post-contact behavior.

Aerial carrier recovery adds those effects. Wang \emph{et al.} studied autonomous takeoff and landing of a small multirotor on a larger six-rotor platform using vision and an RBF-PID controller, demonstrating landing feasibility on an aerial carrier~\cite{wang2022dual}. Shen \emph{et al.} used convex trajectory optimization to plan quadrotor landing trajectories on aerial vehicle carriers while respecting kinodynamic constraints~\cite{shen2024convex}. Dong \emph{et al.} introduced an aerodynamic-interference region and landing cone, and validated real-time trajectory optimization for both hovering and moving carriers; their moving-carrier field test used a predetermined carrier speed of \SI{0.5}{m/s}, with OptiTrack supplying both UAV poses, reporting approximately \SI{13}{cm} dynamic landing accuracy over three trials~\cite{dong2024aerial}. Recent work has demonstrated compound-VTOL docking using RTK rendezvous, monocular relative positioning, and PID control~\cite{10851574}. A recurring assumption across these studies is a passive recovery platform with poses supplied by off-board infrastructure; the present system instead closes the full carried-deployment loop with fully onboard sensing. Previous work on aerial mother--child recovery has investigated field-validated recovery onto a hovering carrier using model-based guidance, disturbance rejection, feasibility constraints, and
safety filtering. In contrast, the present work investigates a
learned RTK--vision recovery policy and explicitly considers recovery
after the carrier translates from the deployment location. \cite{mother29} The
experiments reported here constitute a separate evaluation campaign
focused on the learned moving-carrier recovery architecture.

\textbf{Comparison with this work.} Table~\ref{tab:positioning} positions the present system against these representative works. Whereas prior aerial-carrier works rely on external motion capture, stored launch coordinates, or pure classical control, the present system combines onboard RTK with fiducial-marker vision, employs a learned terminal recovery policy, and closes the full operational loop including carrier deployment, independent sortie, and verified re-recovery. The PPO policy achieves 99.55\% success against 78.4\% for a tuned PD baseline under identical simulation evaluation conditions.

\subsection{Fiducial marker detection for precision landing}

Precision landing pipelines commonly rely on engineered fiducial markers to reduce ambiguity and improve detectability during the final approach meters~\cite{olson2011apriltag,garrido2014aruco,kalaitzakis2021fiducial}. ArUco and AprilTag families provide robust, low-latency 6-DoF pose estimation from a single monocular image and have been employed across UAV landing~\cite{baca2019autonomous,semerikov2025vision}, docking onto moving ground platforms~\cite{lange2009autonomous,baca2019autonomous}, and aerial-target rendezvous~\cite{wang2022dual}. The present system uses a QR-code fiducial on the recovery deck, detected by an OpenMV camera; the detection pipeline provides deck-relative $x$-$y$ offset at up to \SI{30}{fps}, fused with RTK in the terminal phase.

\subsection{Learning-based flight and sim-to-real transfer}

Reinforcement learning has been used for quadrotor stabilization~\cite{hwangbo2017}, moving-platform landing~\cite{rodriguezramos2019}, and aggressive flight~\cite{kaufmann2023}. Neural Lander showed that learned dynamics can compensate near-surface aerodynamic effects~\cite{shi2019neurallander}, and domain randomization is widely used to reduce the simulation-to-reality gap~\cite{tobin2017domain}. The present work uses PPO~\cite{schulman2017ppo} and a physics-based MuJoCo transfer model~\cite{todorov2012mujoco}, with explicit sensor noise, aerodynamic disturbance, and marker-latency randomization, embedding the learned recovery behavior inside a full carrier-deployment mission with a moving airborne recovery target.

\begin{table}[t]
\caption{Positioning relative to representative aerial-carrier landing studies. Success rates are evaluated under each work's own conditions and are not directly comparable across works without a matched-baseline campaign.}
\label{tab:positioning}
\centering
\scriptsize
\setlength{\tabcolsep}{2.5pt}
\begin{tabular}{p{0.15\columnwidth}p{0.19\columnwidth}p{0.19\columnwidth}p{0.18\columnwidth}p{0.16\columnwidth}}
\toprule
Work & Relative sensing & Recovery method & Mission emphasis & Reported accuracy \\
\midrule
Wang \emph{et al.} \cite{wang2022dual} & Onboard vision & RBF-PID & Aerial takeoff/landing & cm-level \\
Shen \emph{et al.} \cite{shen2024convex} & Carrier-state based & Convex opt.\ & Aerial-carrier landing & Feasible traj.\ \\
Dong \emph{et al.} \cite{dong2024aerial} & External OptiTrack & Interference-aware opt.\ & Hovering/moving landing & $\approx$13 cm (3 trials) \\
\textbf{This work} & \textbf{Shared RTK + fiducial vision + range} & \textbf{PPO + safety gates} & \textbf{Deploy, sortie, recover, descend} & \textbf{6.62 cm median (2000 ep.)} \\
\bottomrule
\end{tabular}
\end{table}

\section{System Architecture and Autonomous Mission}
\begin{figure*}[t]
    \centering
    \includegraphics[width=0.98\textwidth]{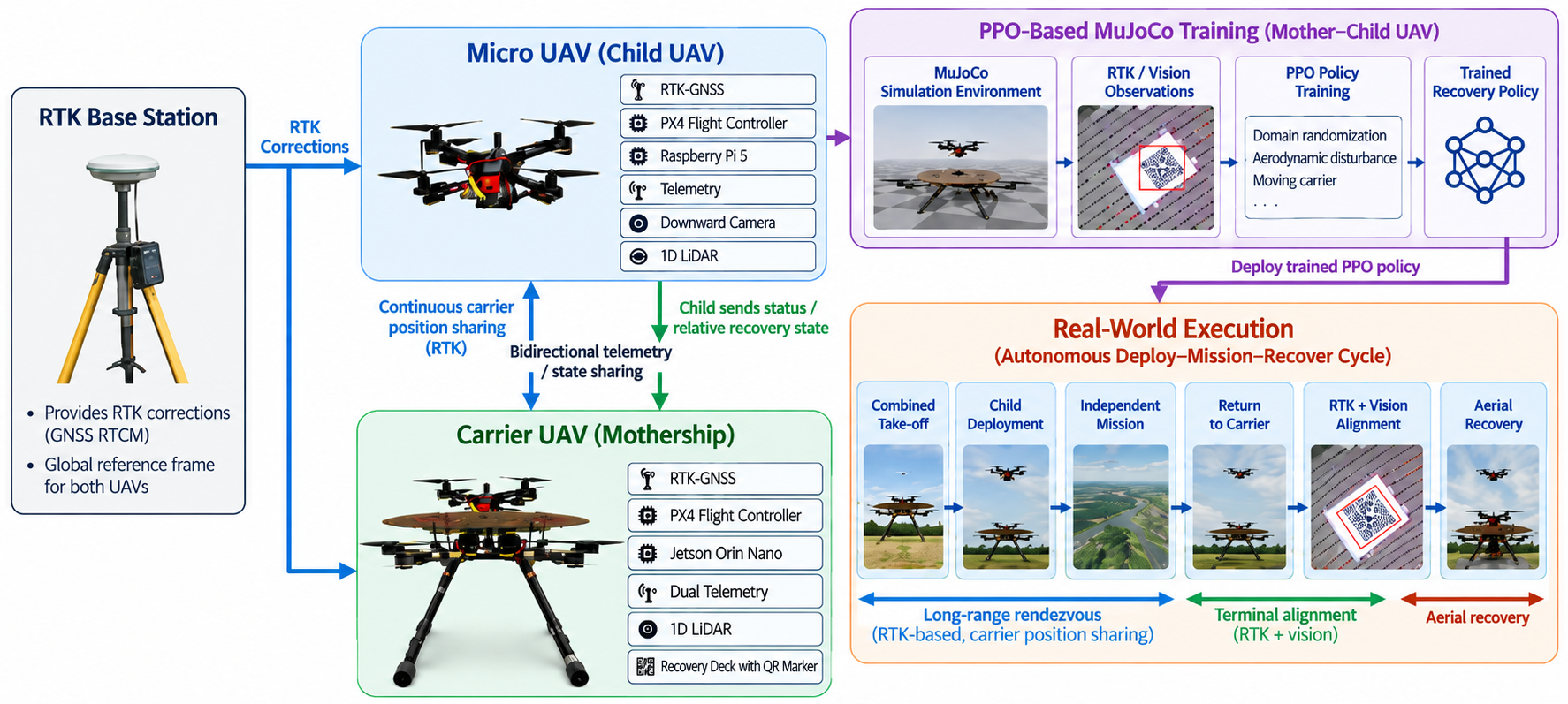}
    \caption{System architecture and autonomous execution flow. Both UAVs receive RTK corrections and maintain independent RTK-GNSS states. The carrier continuously shares its current position with the child through bidirectional telemetry. The child uses RTK for long-range rendezvous and adds downward-camera fiducial marker observations in the terminal region. The learned recovery policy is deployed at the navigation layer, while PX4 provides low-level flight control. Safety gate logic provides a deterministic, policy-independent descent authorization layer, and dual onboard range/state checks confirm recovery.}
    \label{fig:architecture}
\end{figure*}

\begin{figure}[t]
    \centering
    \includegraphics[width=0.8\columnwidth]{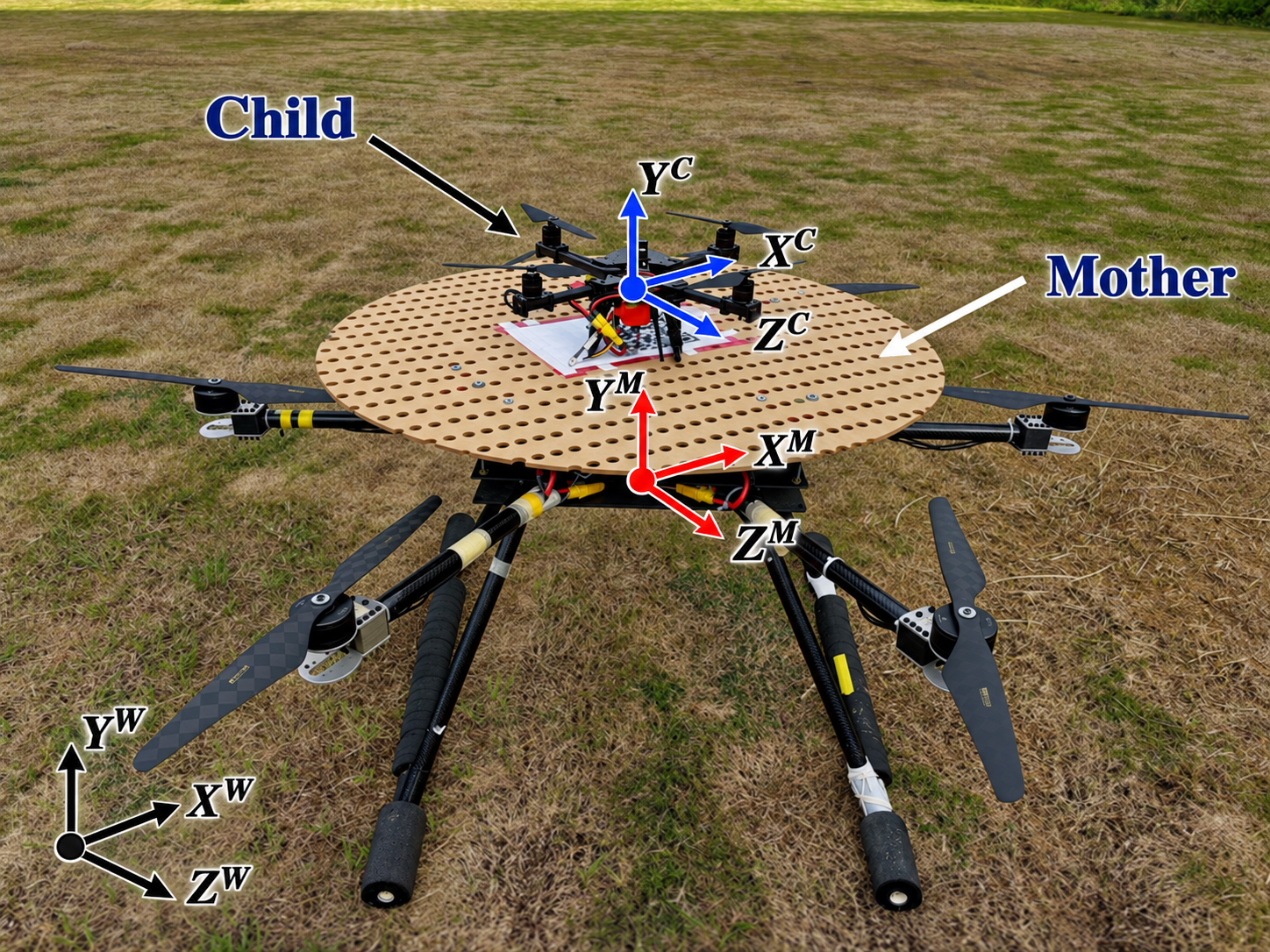}
    \caption{Physical mother--child platform and reference-frame definition. The inertial frame is $\mathcal F_W=\{X^W,Y^W,Z^W\}$, the carrier body frame is $\mathcal F_M=\{X^M,Y^M,Z^M\}$, and the child body frame is $\mathcal F_C=\{X^C,Y^C,Z^C\}$.}
    \label{fig:reference_frames}
\end{figure}

\subsection{Mother--child hardware}

The carrier is a \MotherMass~kg hexarotor using six \SI{360}{KV} motors with 22$\times$6~in propellers. A circular \DeckDia~m-diameter recovery plate is mounted above the carrier rotor plane and carries the fiducial landing marker. The child is a \ChildMass~kg quadrotor of approximately \ChildDia~m overall diameter using T-Motor AIR~2216 \SI{920}{KV} motors and 10$\times$4.5~in propellers. The recovered nominal stack mass is therefore approximately \SI{15.9}{kg}. Both vehicles use PX4 for low-level flight control and carry independent RTK-GNSS receivers. The mothership carries an NVIDIA Jetson Orin Nano Super, a one-dimensional range sensor, ATK telemetry, and a secondary supervisory controller. The child carries a Raspberry Pi~5, a downward-facing OpenMV camera running a QR-code fiducial detector at up to \SI{30}{fps}, a one-dimensional range sensor, ATK telemetry, and a secondary supervisory controller. Hardware specifications are summarized in Table~\ref{tab:hardware}.

\begin{table}[t]
\caption{Physical platform used in the experiments.}
\label{tab:hardware}
\centering
\scriptsize
\setlength{\tabcolsep}{4pt}
\begin{tabular}{lll}
\toprule
Item & Carrier & Child \\
\midrule
Configuration & Hexarotor & Quadrotor \\
Mass & \MotherMass~kg & \ChildMass~kg \\
Propulsion & 6$\times$360 KV & 4$\times$920 KV \\
Propellers & 22$\times$6 in & 10$\times$4.5 in \\
Recovery/vehicle dia. & \DeckDia~m deck & $\approx$\ChildDia~m \\
Companion compute & Jetson Orin Nano Super & Raspberry Pi~5 \\
Global navigation & RTK-GNSS & RTK-GNSS \\
Near-field sensing & 1-D range & OpenMV + 1-D range \\
Marker detector & -- & QR fiducial, $\leq$30 fps \\
\bottomrule
\end{tabular}
\end{table}

\subsection{Continuous RTK return and vision-aided alignment}

Let $\mathbf p_M^W$ and $\mathbf p_C^W$ denote the carrier and child positions in the common navigation frame. The long-range vector from the child to the current carrier position is
\begin{equation}
\mathbf e_{\rm RTK}=\mathbf p_M^W-\mathbf p_C^W,
\label{eq:rtkrel}
\end{equation}
with relative velocity $\mathbf v_r=\mathbf v_M^W-\mathbf v_C^W$. Because the carrier broadcasts its state throughout the sortie, the return target is continuously updated. This is essential in the translating-carrier experiment, where returning to the release coordinate would produce an obsolete target.

RTK remains active in the terminal phase. When the recovery marker is detected by the downward camera, the controller additionally obtains a deck-relative visual cue, a visibility flag $s_V$, and the age $\tau_V$ of the last accepted observation. The system thus uses overlapping sensing rather than a hard ``GPS-off/vision-on'' switch: RTK supplies persistent global relative information while the fiducial marker refines alignment to the physical deck.

\subsection{Autonomous mission and docking verification}

The operational state sequence is: (1) carrier takeoff, (2) airborne child release, (3) independent child mission, (4) RTK-guided return, (5) RTK--vision terminal recovery, (6) docking verification, and (7) recovered-stack descent. The child-side range measurement and terminal state generate a local confirmation $q_C$; independently, the carrier uses its range measurement together with the shared child state to generate $q_M$. Recovery is accepted only when
\begin{equation}
q_{\rm dock}=q_C\land q_M,
\label{eq:verify}
\end{equation}
and the condition persists for a short dwell interval ($\geq \SI{0.5}{s}$). Only then does the carrier initiate coupled descent.

\section{RTK--Vision-Guided Reinforcement Learning}
\subsection{Observation and action}

The policy operates at the recovery/navigation layer rather than commanding motors directly; PX4 retains low-level stabilization~\cite{meier2015px4}. It governs only the terminal recovery phase of the child, from marker acquisition to capture, and does not command the carrier. A 12-dimensional observation is used,
\begin{equation}
\mathbf o_t=[\hat{\mathbf e}^{T},\mathbf v_C^{T},\tau_V,s_V,\hat v_{M,x},\hat v_{M,y},t_n,a_{z,t-1}]^{T}\in\mathbb R^{12},
\label{eq:obs}
\end{equation}
where $\hat{\mathbf e}$ is the current carrier-relative position estimate, $\mathbf v_C$ is the child velocity, $t_n$ is normalized recovery time, and $a_{z,t-1}$ is the previous vertical action. On hardware, $\hat{\mathbf e}$ is initialized and maintained from the two RTK states and refined near the deck by the onboard fiducial observation. When a new marker observation is unavailable, its last valid contribution is retained while $s_V=0$ and $\tau_V$ increases so that staleness is explicit to the policy.

The actor outputs a normalized three-axis carrier-relative velocity command,
\begin{align}
\mathbf a_t&=[a_x,a_y,a_z]^T\in[-1,1]^3,\\
\mathbf v_{r,t}^{d}&=\operatorname{diag}(1.5,1.5,1.0)\mathbf a_t~\si{m/s}.
\label{eq:action}
\end{align}
The command is mapped to the child navigation reference and tracked by PX4. The safety gate layer (Section~\ref{sec:safety_gate}) may pause unsafe descent and authorize the docking transition independently of the learned horizontal recovery behavior.

\begin{figure}[t]
    \centering
    \includegraphics[width=0.8\columnwidth]{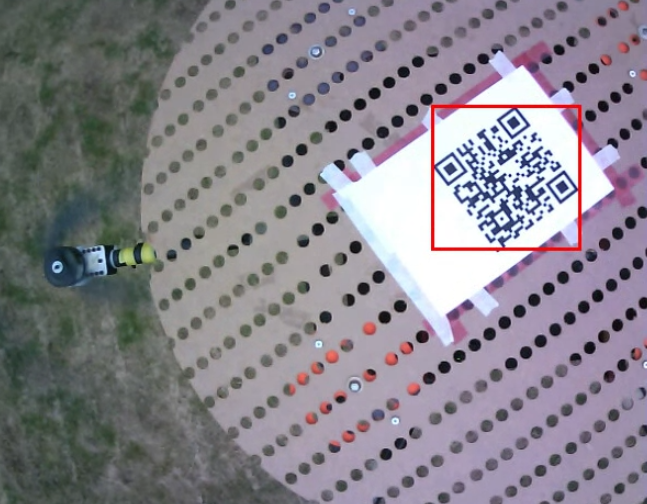}
    \caption{Representative child-camera frame during terminal recovery. The fiducial marker on the carrier deck is detected inside the red bounding region and provides a close-range deck-relative cue while RTK carrier tracking remains active.}
    \label{fig:qr_detection}
\end{figure}

\subsection{Simulation environment and sensor models}
\label{sec:sim_env}

All training and policy transfer take place inside a MuJoCo physics environment~\cite{todorov2012mujoco}. The simulation runs at a physics timestep of \SI{4}{ms} with control decisions at \SI{50}{ms} (\SI{20}{Hz}), matching the onboard navigation loop frequency. Both mother and child are modeled as rigid bodies with six degrees of freedom each; individual motor thrust and torque are actuated explicitly using quadratic thrust models $f_i = k_f \omega_i^2$.

\textbf{Sensor noise models.} RTK-GNSS position noise is modeled as zero-mean Gaussian with standard deviation \SI{2}{cm} horizontal and \SI{4}{cm} vertical per vehicle, consistent with typical RTK horizontal accuracy. The fiducial marker detector introduces discrete latency of 1--8 control steps (\SIrange{50}{400}{ms}) and per-frame dropout probability 0.15, representing motion blur and partial occlusion near the deck edge. Marker $x$-$y$ measurements include additive Gaussian noise ($\sigma=\SI{1.5}{cm}$) and a slow-drifting bias random walk ($\sigma_{\rm walk}=\SI{0.5}{cm\,s^{-1}}$). The 1-D range sensor is modeled with \SI{2}{cm} Gaussian noise and \SI{4}{\percent} random dropout. Camera field-of-view geometry is simulated explicitly so that the marker exits the image when the child drifts beyond the detection cone.

\textbf{Aerodynamic disturbance surrogate.} A compact carrier-wake model uses planar child-to-deck radius $r$ and positive vertical gap $h$,
\begin{equation}
s(r,h)=k_{dw}\exp\!\left(-\frac{r^2}{2\sigma_r^2}\right)\exp\!\left(-\frac{h}{h_0}\right),
\label{eq:wake}
\end{equation}
with parameters $\sigma_r = \SI{0.40}{m}$, $h_0 = \SI{0.60}{m}$. The disturbance gain $k_{dw}$ is randomized uniformly in $[0.5, 2.0]\,\si{N}$ to represent the range of proximity effects reported in aerial-interaction studies~\cite{jain2019aero,dong2024aerial}. Gust disturbances are added as independent Ornstein--Uhlenbeck processes on each vehicle, with standard deviation up to \SI{1.5}{N} at full curriculum difficulty.

\textbf{Transfer model.} The full MuJoCo rigid-body dynamics for vehicle $i\in\{M,C\}$ are:
\begin{align}
 m_i\dot{\mathbf v}_i &= m_i\mathbf g + \mathbf f_{T,i}+\mathbf f_{D,i}+\mathbf f_{W,i}+\mathbf f_{c,i}, \\
 \mathbf J_i\dot{\boldsymbol\omega}_i &= \boldsymbol\tau_i-\boldsymbol\omega_i\times(\mathbf J_i\boldsymbol\omega_i)+\boldsymbol\tau_{c,i},
\label{eq:dynamics}
\end{align}
where $\mathbf f_T$ is rotor thrust, $\mathbf f_D$ is drag, $\mathbf f_W$ groups wake and gust disturbances, and $\mathbf f_c,\boldsymbol\tau_c$ are contact terms. The model includes the recovery deck collision geometry, child landing-leg contact, and a post-capture holonomic constraint.

\subsection{PPO training and randomized terminal environment}

PPO is on-policy, so no fixed trajectory dataset is required. The actor has two 64-unit $\tanh$ layers and a three-dimensional Gaussian action head. To avoid an uninformative sparse-reward start, the actor is warm-started for 350 supervised updates from a stabilizing velocity-feedback teacher; the reported learned checkpoint is then optimized with PPO for 24 updates using 128 parallel environments and 128-step rollouts, yielding 393,216 PPO transitions. The implementation uses four PPO epochs per update, minibatches of 2048, learning rate $8\times10^{-5}$, discount $\gamma=0.992$, GAE parameter $\lambda=0.95$~\cite{schulman2016gae}, and clipping parameter 0.18. A curriculum increases the initial-condition spread and disturbance level over approximately the first 45\% of PPO updates.

At full curriculum difficulty, the child begins at a planar radius of \SIrange{0.30}{1.80}{m} and height of \SIrange{1.50}{3.50}{m}; the initial velocity standard deviation is \SI{0.15}{m/s}, deck velocity is sampled up to \SI{0.50}{m/s} per horizontal axis, and response time constant is sampled from \SIrange{0.16}{0.32}{s}. The trainer also randomizes carrier acceleration, gust acceleration, relative-state bias, camera visibility, and marker measurement noise as described in Section~\ref{sec:sim_env}.

With $r_{xy}=\|\mathbf e_{xy}\|$, horizontal relative speed $v_{xy}$, height error $e_z$, and planar closing rate $\rho$, the reward is
\begin{equation}
\begin{split}
r_t={}&-1.50r_{xy}-0.32v_{xy}-0.30|e_z|-0.018\|\mathbf a_t\|^2\\
&+0.30\operatorname{clip}(\rho,-1,1)+0.08s_V+35I_{\rm succ},
\end{split}
\label{eq:reward}
\end{equation}
where $I_{\rm succ}=1$ when $r_{xy}<\SI{0.16}{m}$, $v_{xy}<\SI{0.28}{m/s}$, the child is within the seated-height corridor, and $|v_z|<\SI{0.35}{m/s}$. This favors low-energy capture rather than geometric coincidence alone.

\subsection{Safety gate control}
\label{sec:safety_gate}

A deterministic safety gate layer, independent of the PPO policy, supervises all descent transitions. The gate evaluates four conditions at each navigation loop step:

\begin{enumerate}
\item \textbf{Lateral alignment:} $r_{xy} < r_{\rm gate}$, where $r_{\rm gate} = \SI{0.20}{m}$.
\item \textbf{Estimate freshness:} marker staleness $\tau_V < \tau_{\rm max} = \SI{0.60}{s}$; if exceeded, descent is paused and the child executes a holding hover until the marker is re-acquired.
\item \textbf{Relative tilt:} the carrier attitude estimate satisfies $|\phi_M|, |\theta_M| < \SI{8}{\degree}$, ensuring the deck is approximately level for contact.
\item \textbf{Relative speed:} vertical approach speed $|v_z| < \SI{0.35}{m/s}$ and lateral speed $v_{xy} < \SI{0.30}{m/s}$.
\end{enumerate}

The gate must pass all four conditions continuously for $t_{\rm dwell} \geq \SI{0.3}{s}$ before descent is authorized. If any condition fails during descent, descent is immediately paused and the child hovers until all conditions recover. This architecture separates the learned lateral alignment behavior from safety-critical descent supervision, reducing the consequence of policy uncertainty near the deck.

In the rigid-body transfer implementation, the terminal lateral command is RL-dominant (85\% PPO, 15\% stabilizing visual-servo), while vertical descent is entirely safety-gate supervised. The field deployment uses the same high-level RTK--vision recovery semantics while PX4 closes the inner-loop dynamics.

\begin{figure*}[t]
    \centering
    \includegraphics[width=0.96\textwidth]{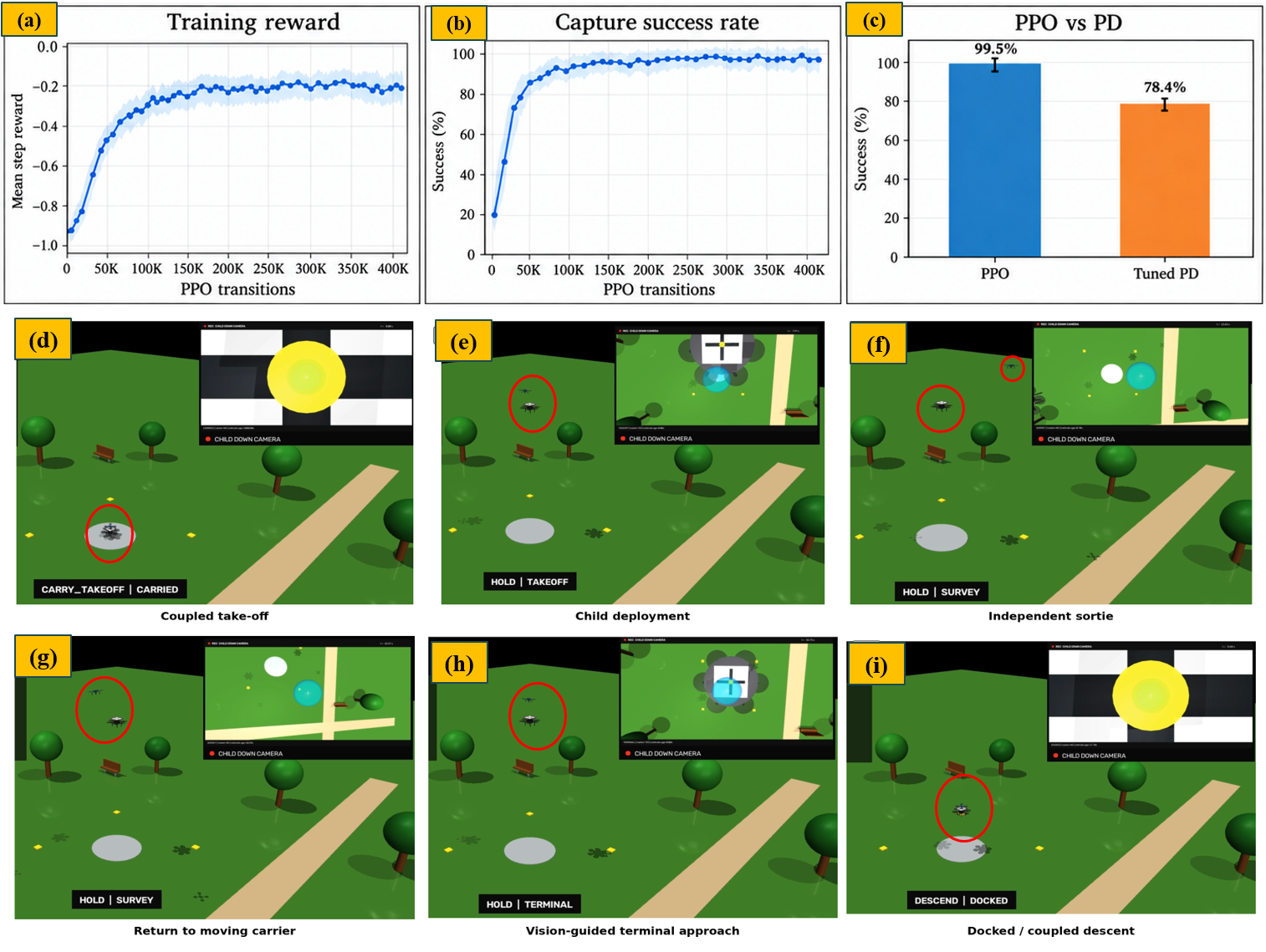}
    \caption{PPO training evidence and physics-based MuJoCo mission transfer. (a)~Mean step reward over the 393,216-transition PPO run; the dashed line marks the transition to full curriculum difficulty. (b)~Deterministic checkpoint success during training, evaluated on independent randomized terminal episodes. (c)~Held-out evaluation over 2,000 episodes: PPO achieves 99.55\% success and \SI{6.62}{cm} median planar terminal error, versus 78.4\% for a tuned PD baseline under identical conditions. (d)--(i)~Representative MuJoCo sequence: coupled takeoff, child deployment, independent sortie, return to the moving carrier, vision-guided terminal approach, and successful docking/coupled descent. Insets show the synchronized downward child camera.}
    \label{fig:rl_mujoco_results}
\end{figure*}

\section{Experimental Evaluation and Results}
\label{sec:experimental_results}

\subsection{Outdoor Protocol}

The proposed deploy--mission--recover framework was evaluated outdoors under two carrier-motion conditions. In both cases, the mother--child system takes off in the coupled configuration, the child UAV departs from the airborne carrier, performs an independent sortie, returns using the continuously shared carrier RTK state, enters the RTK--vision terminal recovery phase, and redocks before the recovered system descends. Seven trials were conducted for each condition. Experiment~1 achieved 7/7 successful missions, whereas Experiment~2 achieved 6/7, giving an overall success rate of 13/14 (\SI{92.9}{\percent}). PX4 flight logs (\texttt{.ulg}) are synchronized using the embedded GNSS UTC timestamp, and horizontal positions are transformed into a common local tangent frame (North-East-Down) using the WGS-84 reference; each vehicle's vertical state is independently ground-referenced before expressing altitude above ground level (AGL), avoiding direct subtraction of local $z$ states with different estimator origins.

\subsection{Experiment~1: Large-Baseline Return}

Experiment~1 represents recovery while the carrier remains within approximately the same local operating region. In Fig.~\ref{fig:field_two_experiments} (top), A denotes coupled take-off and B the airborne child-deployment event; the carrier thrust trace shows a clear transient around B, consistent with the abrupt payload change after release. The mothership re-establishes stable flight, climbs through C, and moves toward D, where it maintains the recovery condition. The child returns using the broadcast carrier RTK position and then uses the downward camera together with RTK for terminal alignment. The maximum synchronized horizontal separation is \SI{44.11}{m}; the carrier is approximately \SI{7.6}{m} AGL at deployment, reaches \SI{8.2}{m} AGL post-release, and displaces approximately \SI{1.65}{m} horizontally from deployment to recovery. After redocking, child- and carrier-side range/state checks confirm recovery before coupled descent.

\subsection{Experiment~2: Translating-Carrier Recovery}

\begin{figure*}[t]
    \centering
    \includegraphics[width=0.9\textwidth]{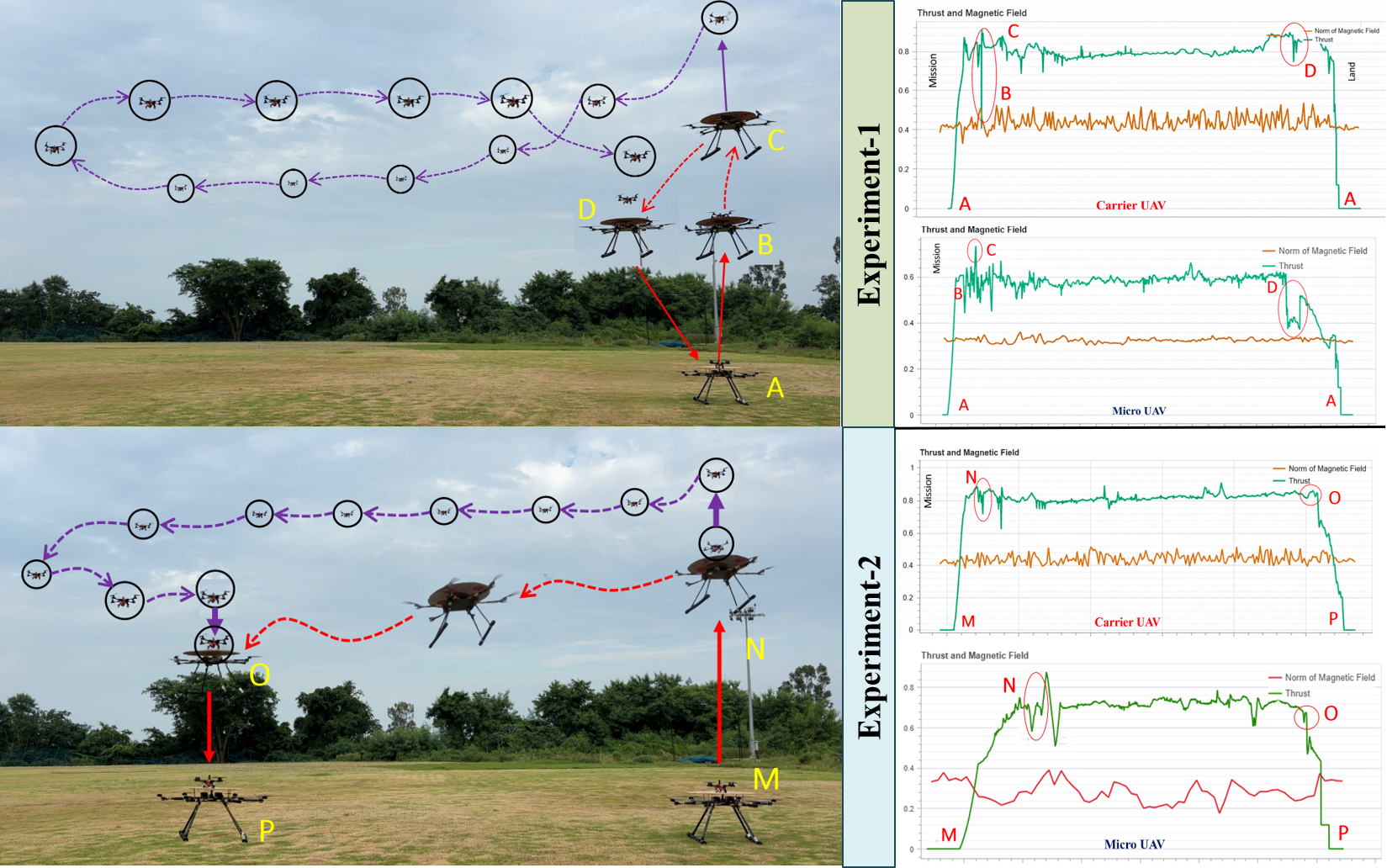}
    \caption{Outdoor autonomous deploy--mission--recover experiments. \textbf{Top, Experiment~1:} the mother--child system starts at A, climbs through B, and reaches child deployment near C. The child executes an independent sortie, returns, and is recovered near D before coupled descent. Right panels show normalized thrust and magnetic-field-norm histories; the carrier thrust signature confirms that the increased stack mass after docking produces a visible thrust step. \textbf{Bottom, Experiment~2:} the system starts at M and the child is deployed near N; the carrier then translates while the child is away. The child returns to the carrier's updated position and is recovered near O before descent to P.}
    \label{fig:field_two_experiments}
\end{figure*}

Experiment~2 evaluates recovery after the mothership changes its airborne position. In Fig.~\ref{fig:field_two_experiments} (bottom), M denotes coupled take-off and N the child-deployment event; after this transition the carrier translates forward rather than remaining at the release position, so the child cannot return to a memorized coordinate and instead tracks the continuously updated carrier RTK state. As the child approaches the current carrier location, the deck marker is detected and the PPO-based RTK--vision policy performs terminal alignment and descent. Frame-by-frame analysis of the \SI{30}{fps} field video identifies child separation at $\approx$\SI{10.85}{s} and first visible docking contact at $\approx$\SI{74.33}{s}, i.e.\ $\approx$\SI{63.48}{s} of independent flight. After docking and verification at O, the recovered stack descends to P. This demonstrates recovery to a translated airborne target, supporting long-endurance missions in which the carrier keeps moving while the child performs a localized task.

\begin{table}[!t]
\centering
\caption{Outdoor evaluation of the deploy--mission--recover framework. Seven trials per carrier-motion condition (14 total). Common to both: fixed RTK, RTK\,+\,onboard marker terminal perception, dual range/state docking verification, and return to the carrier's current position.}
\label{tab:fieldresults}
\scriptsize
\setlength{\tabcolsep}{4pt}
\renewcommand{\arraystretch}{1.1}
\begin{tabular}{p{0.42\columnwidth}p{0.24\columnwidth}p{0.24\columnwidth}}
\toprule
\textbf{Metric} & \textbf{Exp.~1} & \textbf{Exp.~2} \\
\midrule
Recovery mode & Near-region & Translating \\
Successful missions & 7/7 (100\%) & 6/7 (85.7\%) \\
Airborne deployment & 7/7 & 7/7 \\
Return/redocking & 7/7 & 6/7 \\
Recovered-stack descent & 7/7 & 6/7 \\
Independent flight & $\approx$87.1 s & 63.48 s \\
Max.\ horizontal separation & 44.11 m & -- \\
Carrier alt.\ at deployment & $\approx$7.65 m AGL & -- \\
Post-deploy.\ max.\ alt.\ & $\approx$8.28 m AGL & -- \\
\midrule
\textbf{Combined} & \multicolumn{2}{c}{\textbf{13/14 (92.9\%)}} \\
\bottomrule
\end{tabular}
\end{table}

\subsection{Failure case analysis}
\label{sec:failure}

To characterize system failure modes, we analyze both held-out simulation episodes (the 0.45\% failed episodes from the 2,000-episode evaluation) and observations from outdoor trials. Three failure classes were identified.

\textbf{Class~1: Late marker acquisition.} When the initial planar radius exceeds \SI{1.5}{m} and the marker enters view late, the policy inherits stale RTK-only estimates and arrives over the deck at excessive lateral speed ($v_{xy} > \SI{0.45}{m/s}$), triggering the gate lateral-speed condition. Log signatures: $s_V = 0$ for $> \SI{0.4}{s}$ after entering the $r_{xy} < \SI{0.50}{m}$ region, and $\tau_V$ increasing past $\tau_{\rm max}$ before the pause activates. The gate correctly pauses descent in all such cases; the episodes count as failures only because docking did not complete within the time limit.

\textbf{Class~2: Deck tilt exceedance.} When sampled carrier attitude exceeds \SI{7}{\degree}, the tilt gate is intermittently violated, causing pause-and-resume cycles that occasionally prevent landing within the time budget. Log signature: $|\phi_M|$ or $|\theta_M|$ oscillating between 6--9$^{\circ}$ at 2--5 Hz, correlated with reduced gate-dwell completion.

\textbf{Class~3: RTK relative-state bias.} Rare worst-case bias random-walk realizations produce a sustained offset $> \SI{5}{cm}$, stalling the child in a hover at the gate boundary. Log signature: $r_{xy}$ holding near $r_{\rm gate}$ for $> \SI{2.0}{s}$ while $v_{xy} \approx 0$ and $s_V = 1$, indicating the policy is satisfied with an incorrect zero-error estimate.

The one outdoor mission that did not complete (Experiment~2, 1/7) was consistent with Class~1: the child arrived over the deck with high lateral speed following an extended period of marker invisibility due to lighting conditions, the safety gate correctly paused descent, and the trial was terminated conservatively before contact. In every reported trial, no coupled descent was initiated until $q_{\rm dock}$ was satisfied, confirming that the safety layer behaves as designed in the field.

\FloatBarrier
\section{Discussion}

The results support the system-level claim rather than a claim of universal superiority of RL over classical control for this task. Dong~\emph{et al.} demonstrated hovering and moving aerial-carrier landing using interference-aware trajectory optimization and external OptiTrack, with approximately \SI{13}{cm} dynamic accuracy~\cite{dong2024aerial}. The present distinction is the integrated operational loop and onboard sensing: the child is physically carried, deploys from the airborne mother, completes an independent mission, returns to a carrier whose position may have changed, combines shared RTK with onboard fiducial-marker alignment, and is verified onboard before coupled descent. In the held-out evaluation the PPO policy achieves 99.55\% success versus 78.4\% for a tuned PD baseline under identical conditions; a matched-baseline hardware campaign at scale is an important next evaluation.

\textbf{Application areas.} The carrier--child recovery loop enables missions a single micro UAV cannot sustain. In \emph{persistent surveillance and ISR}, the carrier ferries and loiters while the child performs close, agile inspection and rejoins without the carrier landing, extending effective mission time. In \emph{long-transit infrastructure inspection}—power lines, pipelines, rail, wind farms, solar sites—the carrier covers the energy-expensive transit and the child performs close-range inspection. In \emph{disaster response and search-and-rescue}, the architecture combines agile close access with the range of the larger platform, and it applies equally to \emph{last-stretch logistics} from a transiting carrier. Because recovery targets the carrier's live shared state rather than fixed infrastructure, it is also a stepping stone toward GPS-degraded and contested-environment operation, which we identify as future work.

The study has four principal limitations. First, only one child UAV is evaluated; multi-child release, deconfliction, and deck occupancy remain future work. Second, the MuJoCo aerodynamic interaction is a tractable disturbance surrogate rather than blade-resolved CFD, with parameters not yet identified from flight logs. Third, Experiment~2 provides video-derived timing rather than a metrically calibrated carrier trajectory. Fourth, the fiducial detector is demonstrated through recorded onboard detections, but a standalone detection benchmark with quantified pose error is outside the present scope. Future work will address these points together with an active-carrier formulation in which the mother assists recovery, and infrastructure-free relative return for GPS-degraded operation.

\section{Conclusion}

This paper presented an RTK--vision-guided reinforcement-learning framework for autonomous recovery of a micro UAV onto a moving carrier UAV, closing the complete aerial deploy--mission--recover loop with a learned terminal recovery policy, an explicit safety-gate layer, and dual onboard docking verification. The PPO policy, trained in a MuJoCo environment with characterized sensor noise, aerodynamic disturbance, and marker-latency randomization, achieves 99.55\% terminal success and \SI{6.62}{cm} median planar error over 2,000 held-out episodes versus 78.4\% for a tuned PD baseline, and the safety gate prevents unsafe contact in all identified failure cases in both simulation and the field. Across 14 outdoor trials the full mission succeeds in 13 (\SI{92.9}{\percent}), spanning near-same-region and translating-carrier recovery. These results establish a practical foundation for reusable mother--child missions and future multi-child and mobile-logistics operations.

\bibliographystyle{IEEEtran}
\bibliography{ref}

@inproceedings{lee2012,
  author    = {D. Lee and T. Ryan and H. J. Kim},
  title     = {Autonomous landing of a {VTOL} {UAV} on a moving platform using image-based visual servoing},
  booktitle = {Proc.\ IEEE Int.\ Conf.\ Robot.\ Autom.\ (ICRA)},
  year      = {2012},
  pages     = {971--976}
}

@inproceedings{falanga2017,
  author    = {D. Falanga and A. Zanchettin and A. Simovic and J. Delmerico and D. Scaramuzza},
  title     = {Vision-based autonomous quadrotor landing on a moving platform},
  booktitle = {Proc.\ IEEE Int.\ Symp.\ Safety Security Rescue Robot.\ (SSRR)},
  year      = {2017}
}

@article{borowczyk2017,
  author  = {A. Borowczyk and D.-T. Nguyen and A. P.-V. Nguyen and D. Q. Nguyen and D. Saussie and J. L. Ny},
  title   = {Autonomous landing of a multirotor micro air vehicle on a high velocity ground vehicle},
  journal = {IFAC-PapersOnLine},
  volume  = {50},
  number  = {1},
  pages   = {10488--10494},
  year    = {2017}
}

@inproceedings{paris2020,
  author    = {A. Paris and B. T. Lopez and J. P. How},
  title     = {Dynamic landing of an autonomous quadrotor on a moving platform in turbulent wind conditions},
  booktitle = {Proc.\ IEEE Int.\ Conf.\ Robot.\ Autom.\ (ICRA)},
  year      = {2020},
  pages     = {9577--9583}
}

@article{wang2022dual,
  author  = {L. Wang and X. Jiang and D. Wang and L. Wang and Z. Tu and J. Ai},
  title   = {Research on aerial autonomous docking and landing technology of dual multi-rotor {UAV}},
  journal = {Sensors},
  volume  = {22},
  number  = {23},
  pages   = {9066},
  year    = {2022}
}

@article{shen2024convex,
  author  = {Z. Shen and G. Zhou and H. Huang and C. Huang and Y. Wang and F.-Y. Wang},
  title   = {Convex optimization-based trajectory planning for quadrotors landing on aerial vehicle carriers},
  journal = {IEEE Trans.\ Intell.\ Veh.},
  volume  = {9},
  number  = {1},
  pages   = {138--150},
  year    = {2024}
}

@article{dong2024aerial,
  author  = {X. Dong and H. Li and Y. Cui and J. Xiang and D. Li and Z. Tu},
  title   = {Aerial landing of micro {UAVs} on moving platforms considering aerodynamic interference},
  journal = {IEEE Robot.\ Autom.\ Lett.},
  volume  = {9},
  number  = {11},
  pages   = {10089--10096},
  year    = {2024}
}

@INPROCEEDINGS{10851574,
  author={Sahu, Aashish and Kumar, R Prasanth},
  booktitle={2024 9th International Conference on Robotics and Automation Engineering (ICRAE)}, 
  title={Design and Implementation of Hexacopter Drone with Integrated Suction and Lift Mechanism with Real-Time Depth Sensing for Precision Object Handling}, 
  year={2024},
  volume={},
  number={},
  pages={6-11},
  doi={10.1109/ICRAE64368.2024.10851574}}

@article{hwangbo2017,
  author  = {J. Hwangbo and I. Sa and R. Siegwart and M. Hutter},
  title   = {Control of a quadrotor with reinforcement learning},
  journal = {IEEE Robot.\ Autom.\ Lett.},
  volume  = {2},
  number  = {4},
  pages   = {2096--2103},
  year    = {2017}
}

@article{rodriguezramos2019,
  author  = {A. Rodriguez-Ramos and C. Sampedro and H. Bavle and P. de la Puente and P. Campoy},
  title   = {A deep reinforcement learning strategy for {UAV} autonomous landing on a moving platform},
  journal = {J.\ Intell.\ Robot.\ Syst.},
  volume  = {93},
  pages   = {351--366},
  year    = {2019}
}

@article{kaufmann2023,
  author  = {E. Kaufmann and L. Bauersfeld and A. Loquercio and M. M\"{u}ller and V. Koltun and D. Scaramuzza},
  title   = {Champion-level drone racing using deep reinforcement learning},
  journal = {Nature},
  volume  = {620},
  pages   = {982--987},
  year    = {2023}
}

@inproceedings{shi2019neurallander,
  author    = {G. Shi and others},
  title     = {Neural Lander: Stable drone landing control using learned dynamics},
  booktitle = {Proc.\ IEEE Int.\ Conf.\ Robot.\ Autom.\ (ICRA)},
  year      = {2019}
}

@inproceedings{tobin2017domain,
  author    = {J. Tobin and R. Fong and A. Ray and J. Schneider and W. Zaremba and P. Abbeel},
  title     = {Domain randomization for transferring deep neural networks from simulation to the real world},
  booktitle = {Proc.\ IEEE/RSJ Int.\ Conf.\ Intell.\ Robots Syst.\ (IROS)},
  year      = {2017}
}

@misc{schulman2017ppo,
  author = {J. Schulman and F. Wolski and P. Dhariwal and A. Radford and O. Klimov},
  title  = {Proximal policy optimization algorithms},
  year   = {2017},
  eprint = {1707.06347},
  archivePrefix = {arXiv}
}

@inproceedings{schulman2016gae,
  author    = {J. Schulman and P. Moritz and S. Levine and M. Jordan and P. Abbeel},
  title     = {High-dimensional continuous control using generalized advantage estimation},
  booktitle = {Proc.\ Int.\ Conf.\ Learn.\ Represent.\ (ICLR)},
  year      = {2016}
}

@inproceedings{todorov2012mujoco,
  author    = {E. Todorov and T. Erez and Y. Tassa},
  title     = {{MuJoCo}: A physics engine for model-based control},
  booktitle = {Proc.\ IEEE/RSJ Int.\ Conf.\ Intell.\ Robots Syst.\ (IROS)},
  year      = {2012},
  pages     = {5026--5033}
}

@inproceedings{meier2015px4,
  author    = {L. Meier and D. Honegger and M. Pollefeys},
  title     = {{PX4}: A node-based multithreaded open source robotics framework for deeply embedded platforms},
  booktitle = {Proc.\ IEEE Int.\ Conf.\ Robot.\ Autom.\ (ICRA)},
  year      = {2015}
}

@inproceedings{jain2019aero,
  author    = {K. P. Jain and T. Fortm\"{u}ller and J. Byun and S. A. Makiharju and M. W. M\"{u}ller},
  title     = {Modeling of aerodynamic disturbances for proximity flight of multirotors},
  booktitle = {Int.\ Conf.\ Unmanned Aircr.\ Syst.\ (ICUAS)},
  year      = {2019},
  pages     = {1261--1269}
}

@inproceedings{olson2011apriltag,
  author    = {E. Olson},
  title     = {{AprilTag}: A robust and flexible visual fiducial system},
  booktitle = {Proc.\ IEEE Int.\ Conf.\ Robot.\ Autom.\ (ICRA)},
  year      = {2011},
  pages     = {3400--3407}
}

@article{garrido2014aruco,
  author  = {S. Garrido-Jurado and R. Mu{\~n}oz-Salinas and F. J. Madrid-Cuevas and M. J. Mar\'{i}n-Jim\'{e}nez},
  title   = {Automatic generation and detection of highly reliable fiducial markers under occlusion},
  journal = {Pattern Recognit.},
  volume  = {47},
  number  = {6},
  pages   = {2280--2292},
  year    = {2014}
}

@article{kalaitzakis2021fiducial,
  author  = {M. Kalaitzakis and B. Cain and S. Carroll and A. Ambrosi and C. Whitehead and N. Vitzilaios},
  title   = {Fiducial markers for pose estimation},
  journal = {J.\ Intell.\ Robot.\ Syst.},
  volume  = {101},
  pages   = {71},
  year    = {2021}
}

@article{baca2019autonomous,
  title={Autonomous landing on a moving vehicle with an unmanned aerial vehicle},
  author={Baca, Tomas and Stepan, Petr and Spurny, Vojtech and Hert, Daniel and Penicka, Robert and Saska, Martin and Thomas, Justin and Loianno, Giuseppe and Kumar, Vijay},
  journal={Journal of Field Robotics},
  volume={36},
  number={5},
  pages={874--891},
  year={2019},
  publisher={Wiley Online Library}
}

@article{semerikov2025vision,
  title={Vision-based autonomous UAV landing: A comprehensive review of technologies, techniques, and applications},
  author={Semerikov, Serhiy O and Nechypurenko, Pavlo P and Vakaliuk, Tetiana A and Mintii, Iryna S and Kolhatin, Andrii O},
  journal={Journal of Intelligent \& Robotic Systems},
  volume={111},
  number={4},
  pages={115},
  year={2025},
  publisher={Springer}
}

@inproceedings{lange2009autonomous,
  author    = {S. Lange and N. Sünderhauf and P. Protzel},
  title     = {Autonomous landing for a multirotor {UAV} using visual pose estimation},
  booktitle = {Proc.\ Eur.\ Conf.\ Mobile Robots (ECMR)},
  year      = {2009}
}

@article{mother29,
  title   = {Integrated Guidance and Control of a Mother--Child UAV--UGV System for Cooperative Missions},
  journal = {arXiv preprint},
  year    = {2026}
}

\end{document}